\def\year{2019}\relax
\documentclass[letterpaper]{article} 
\usepackage{aaai19}  
\usepackage{times}  
\usepackage{helvet}  
\usepackage{courier}  
\usepackage{url}  
\usepackage{graphicx}  
\usepackage{amsmath}
\usepackage{amssymb}
\usepackage{makecell,multirow,diagbox}
\usepackage{subcaption}
\usepackage{bm}
\usepackage{array}
\usepackage[pagebackref=true,breaklinks=true,letterpaper=true,colorlinks,bookmarks=false]{hyperref}
\usepackage{booktabs}
\usepackage{enumerate}
\usepackage[hyphenbreaks]{breakurl}

\begin{document}
%
\title{TrafficPredict: Trajectory Prediction for Heterogeneous Traffic-Agents}
\author{Yuexin Ma$^{1,2}$, Xinge Zhu$^{3}$, Sibo Zhang$^{1}$, Ruigang Yang$^{1}$, Wenping Wang$^{2}$, Dinesh Manocha$^{4}$\\
Baidu Research, Baidu Inc.$^{1}$, The University of Hong Kong$^{2}$,\\
The Chinese University of Hong Kong$^{3}$, University of Maryland at College Park$^{4}$\\
yxma@cs.hku.hk, zhuxinge123@gmail.com, sibozhang@baidu.com, \\
yangruigang@baidu.com, wenping@cs.hku.hk, dm@cs.umd.edu\\
}

\maketitle
\begin{abstract}
To safely and efficiently navigate in complex urban traffic, autonomous vehicles must make responsible predictions in relation to surrounding traffic-agents (vehicles, bicycles, pedestrians, etc.). A challenging and critical task is to explore the movement patterns of different traffic-agents and predict their future trajectories accurately to help the autonomous vehicle make reasonable navigation decision. To solve this problem, we propose a long short-term memory-based (LSTM-based) realtime traffic prediction algorithm, TrafficPredict. Our approach uses an instance layer to learn instances' movements and interactions and has a category layer to learn the similarities of instances belonging to the same type to refine the prediction. In order to evaluate its performance, we collected trajectory datasets in a large city consisting of varying conditions and traffic densities. The dataset includes many challenging scenarios where vehicles, bicycles, and pedestrians move among one another. We evaluate the performance of TrafficPredict on our new dataset and highlight its higher accuracy for trajectory prediction by comparing with prior prediction methods. 
\end{abstract}

\section{Introduction}
Autonomous driving is a significant and difficult task that has the potential to impact people’s day-to-day lives. The goal is to make a vehicle perceive the environment and safely and efficiently navigate any traffic situation without human intervention. Some of the challenges arise in dense urban environments, where the traffic consists of different kinds of {\em traffic-agents}, including cars, bicycles, buses, pedestrians, etc.. These traffic-agents have different shapes, dynamics, and varying behaviors and can be regarded as an instance of a heterogeneous multi-agent system. To guarantee the safety of autonomous driving, the system should be able to analyze the motion patterns of other traffic-agents and predict their future trajectories so that the autonomous vehicle can make appropriate navigation decisions. 

Driving in an urban environment is much more challenging than driving on a highway. Urban traffic is riddled with more uncertainties, complex road conditions, and diverse traffic-agents, especially on some cross-roads. Different traffic-agents have different motion patterns. At the same time, traffic-agents’ behaviors are deeply affected by other traffic-agents. It is necessary to consider the interaction between the agent to improve the accuracy of trajectory prediction.

\begin{figure}
\includegraphics[width=1\columnwidth]{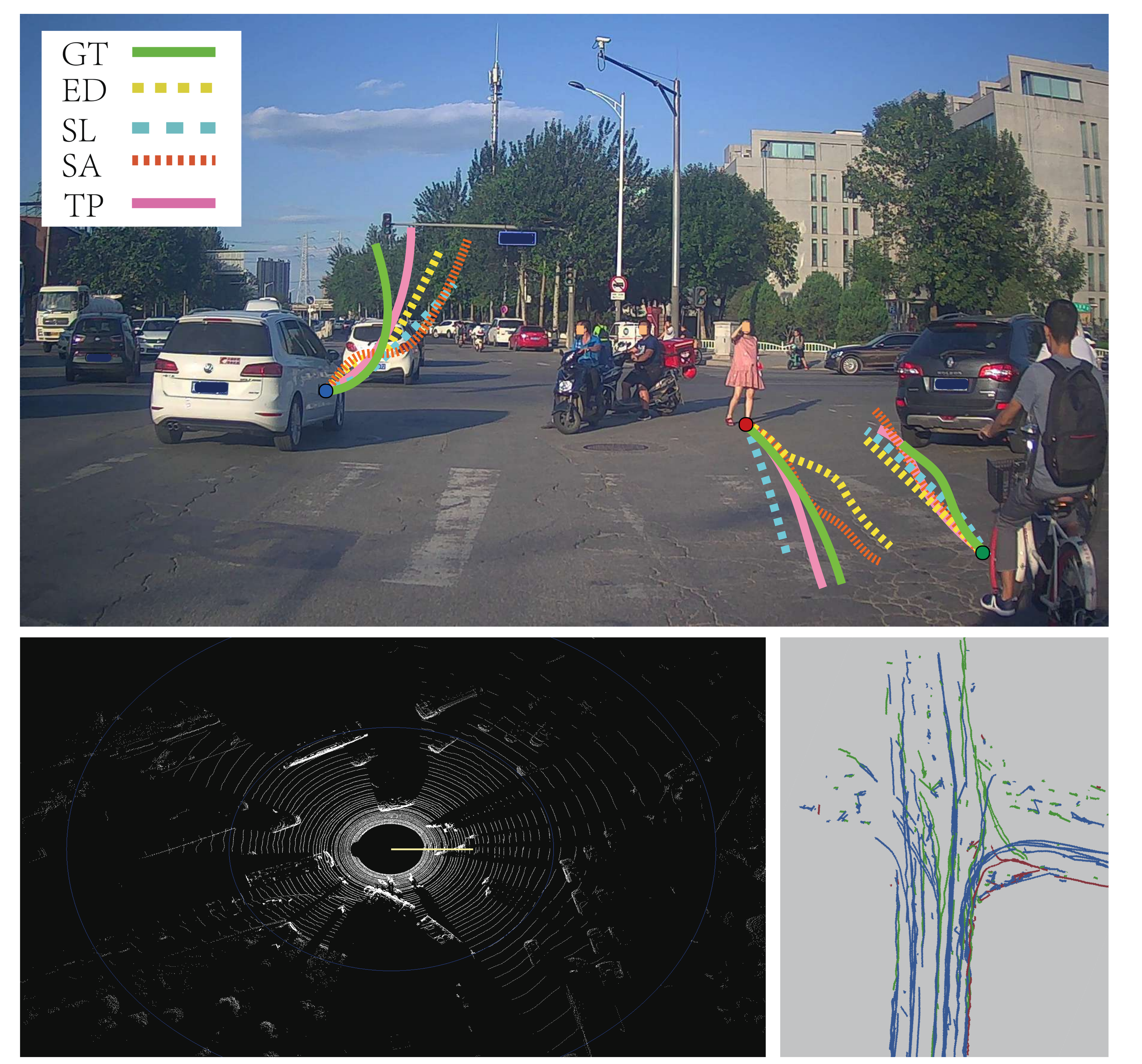}
\caption{{\bf Heterogeneous urban traffic scenario:} We demonstrate the improved trajectory prediction accuracy of our method over prior approaches (top). The  green solid lines denote ground truth trajectories (GT), pink solid lines  are for our method (TP) and dashed lines are the predicted trajectories for other methods (ED, SL, SA). We observe 20\% improvement in accuracy using TP. Traffic corresponding point cloud captured by LiDAR of the acquisition car is shown on the left bottom. Original trajectories of traffic-agents in the scenario are shown on the right bottom: blue for cars, green for bicycles, and red for pedestrians.}
\label{fig:traffic_sample}
\vspace{-4ex}
\end{figure}

The problem of predicting trajectories for moving agents has been studied extensively. Some traditional algorithms are based on motion models like kinematic and dynamic models~\cite{toledo2009imm}, Bayesian filters~\cite{kalman1960new}, Gaussian Processes~\cite{rasmussen2006gaussian}, etc. These methods do not take into account interactions between the traffic-agents and the environment, making it difficult to analyze complicated scenarios or perform long-term predictions. With the success of LSTM networks in modeling non-linear temporal dependencies~\cite{ma2017image} in sequence learning and generation tasks, more and more works have been using these networks to predict trajectories of human crowds\cite{alahi2016social} and vehicles trajectories~\cite{lee2017desire}. The common limitation of these works is the focus on predicting one type of group (only pedestrians or cars, for example). These methods may not work in heterogeneous traffic, where different vehicles and pedestrians coexist and interact with each other\cite{DeepTAgent2018}.

\noindent {\bf Main Results:} For the task of trajectory prediction in heterogeneous traffic, we propose a novel LSTM-based algorithm, TrafficPredict. Given a sequence of trajectory data, we construct a \textit{4D Graph}, where two dimensions are for instances and their interactions, one dimension is for time series, and one dimension is for high-level categorization. In this graph, all the valid instances and categories of traffic-agents are denoted as nodes, and all the relationships in spatial and temporal space is represented as edges. Sequential movement information and interaction information are stored and transferred by these nodes and edges. Our LSTM network architecture is constructed on the \textit{4D Graph}, which can be divided into two main layers: one is the instance layer and the other is the category layer. The former is designed to capture dynamic properties and and interactions between the traffic-agents at a micro level. The latter aims to conclude the behavior similarities of instances of the same category using a macroscopic view and guide the prediction for instances in turn. We also use a self attention mechanism in the category layer to capture the historical movement patterns and highlight the category difference. Our method is the first to integrate the trajectory prediction for different kinds of traffic-agents in one unified framework. 

To better expedite research progress on prediction and navigation in challenging scenarios for autonomous driving, we provide a new trajectory dataset for complex urban traffic with heterogeneous traffic-agents during rush hours. Scenario and data sample of our dataset is shown in Fig.~\ref{fig:traffic_sample}. In practice, TrafficPredict takes about a fraction of a second on a single CPU core and exhibits $20\%$ accuracy improvement over prior prediction schemes. The novel components of our work include:
\begin{itemize}
\item {Propose a new approach for trajectory prediction in heterogeneous traffic.}
\item {Collect a new trajectory dataset in urban traffic with much interaction between different categories of traffic-agents.}
\item {Our method has smaller prediction error compared with other state-of-art approaches.}
\end{itemize}

The rest of the paper is organized as follows. We give a brief overview of related prior work in Section 2. In Section 3, we define the problem and give details of our prediction algorithm. We introduce our new traffic dataset and show the performance of our methods in Section 4.

\section{Related Work}

\subsection{Classical methods for trajectory prediction}
The problem of trajectory prediction or path prediction has been extensively studied. There are many classical approaches, including Bayesian networks~\cite{lefevre2011exploiting}, Monte Carlo Simulation~\cite{danielsson2007monte}, Hidden Markov Models (HMM)~\cite{firl2012predictive}, Kalman
Filters~\cite{kalman1960new}, linear and non-linear Gaussian Process regression models~\cite{rasmussen2006gaussian}, etc. These methods focus on analyzing the inherent regularities of objects themselves based on their previous movements. They can be used in simple traffic scenarios in which there are few interactions among cars, but these methods may not work well when different kinds of vehicles and pedestrians appear at the same time. 

\subsection{Behavior modeling and interactions}
There is considerable work on human behavior and interactions. The Social Force model~\cite{helbing1995social} presents a pedestrian motion model with attractive and repulsive forces, which has been  extended by~\cite{yamaguchi2011you}. Some similar methods have also been proposed that use continuum dynamics~\cite{treuille2006continuum}, Gaussian processes~\cite{wang2008gaussian}, etc. Bera et al.~\shortcite{bera2016glmp,bera2017aggressive} combine an Ensemble Kalman Filter and human motion model to predict the trajectories for crowds. These methods are useful for analyzing motions of pedestrians in different scenarios, such as shopping malls, squares, and pedestrian streets. There are also some approaches to classify group emotions or identify driver behaviors~\cite{cheung2018identifying}. To extend these methods to general traffic scenarios, \cite{ma2018autorvo} predicts the trajectories of multiple traffic-agents by considering kinematic and dynamic constraints. However, this model assumes perfect sensing and shape and dynamics information for all of the traffic agents. 

\begin{figure*}
\includegraphics[width=2.1\columnwidth]{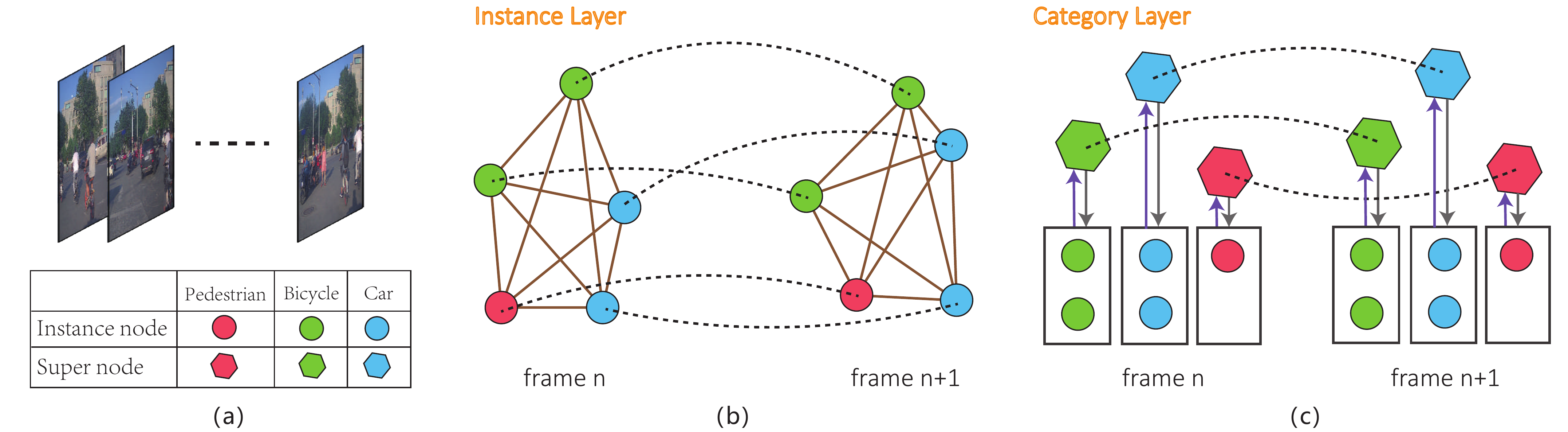}
\caption{Our \textit{4D Graph} for a traffic sequence. (a) Icons for instances and categories are shown on the left table. (b) The instance layer of the \textit{4D Graph} with spatial edges as solid lines and temporal edges as dashed lines. (c) The category layer with temporal edges of super nodes drawn by dashed lines. }
\label{fig:network}
\vspace{-3ex}
\end{figure*}

\subsection{RNN networks for sequence prediction}
In recent years, the concept of the deep neural network (DNN) has received a huge amount of attention due to its good performance in many areas~\cite{goodfellow2016deep}. Recurrent neural network (RNN) is one of the DNN architectures and is widely used for sequence generation in many domains, including speech recognition~\cite{graves2014towards}, machine translation~\cite{chung2015recurrent}, and image captioning~\cite{vinyals2015show}. Many methods based on long short-term Memory (LSTM), one variant of RNN, have been proposed for maneuver classification~\cite{khosroshahi2017learning} and trajectory prediction~\cite{altche2017lstm}. Some methods~\cite{kim2017probabilistic,park2018sequence,lee2017desire} produce the probabilistic information about the future locations of vehicles over an occupancy grid map or samples by making use of an encoder-decoder structure. However, these sampling-based methods suffer from inherent inaccuracies due to discretization limits. Another method~\cite{deo2018multi} presents a model that outputs the multi-modal distribution and then generates trajectories. Nevertheless, most of these methods require clear road lanes and simple driving scenarios without other types of traffic-agents passing through. Based on images, ~\cite{chandra2018traPHic} models the interactions between different traffic-agents by a LSTM-CNN hybrid network for trajectory prediction. Taking into account the human-human interactions, some approaches~\cite{alahi2016social,gupta2018social,vemula2017social} use LSTM for predicting trajectories of pedestrians in a crowd and they show good performance on public crowd datasets. However, these methods are also limited in terms of trajectory prediction in complex traffic scenarios where the interactions are among not only pedestrians but also heterogeneous traffic-agents.

\subsection{Traffic datasets}

There are several datasets related to traffic scenes. Cityscapes ~\cite{cordts2016cityscapes} contains 2D semantic, instance-wise, dense pixel annotations for 30 classes. ApolloScape~\cite{huang2018apolloscape} is a large-scale comprehensive dataset of street views that contains higher scene complexities, 2D/3D annotations and pose information, lane markings and video frames. However these two dataset do not provide trajectories information. The Simulation (NGSIM) dataset ~\cite{USHighway} has trajectory data for cars, but the scene is limited to highways with similar simple road conditions. KITTI ~\cite{geiger2013vision} is a dataset for different computer vision tasks such as stereo, optical ﬂow, 2D/3D object detection, and tracking. However, the total time of the dataset with tracklets is about 22 minutes. In addition, there are few intersection between vehicles, pedestrians and cyclists in KITTI, which makes it insufficient for exploring the motion patterns of traffic-agents in challenging traffic conditions. There are some pedestrian trajectory datasets like ETH ~\cite{pellegrini2009you}, UCY ~\cite{lerner2007crowds}, etc., but such datasets only focus on human crowds without any vehicles.

\section{TrafficPredict}

In this section, we present our novel algorithm to predict the trajectories of different traffic-agents.

\subsection{Problem Definition}

We assume each scene is preprocessed to get the categories and spatial coordinates of traffic-agents. At any time $t$, the feature of the $i$th traffic-agent $A_i^t$ can be denoted as $f_i^t = (x_i^t, y_i^t, c_i^t)$, where the first two items are coordinates in the x-axis and y-axis respectively, and the last item is the category of the traffic-agent. In our dataset, we currently take into account three types of traffic-agents, $c_i\in\{1,2,3\}$, where $1$ stands for pedestrians, $2$ represents bicycles and $3$ denotes cars. Our approach can be easily extended to take into account more agent types. Our task is to observe features of all the traffic-agents in the time interval $[1:T_{obs}]$ and then predict their discrete positions at $[T_{obs}+1: T_{pred}]$. 

\subsection{\textit{4D Graph} Generation}
In urban traffic scenarios where various traffic-agents are interacting with others, each instance has its own state in relation to the interaction with others at any time and they also have continuous information in time series. Considering traffic-agents as instance nodes and relationships as edges, we can construct a graph in the instance layer, shown in Fig.\ref{fig:network} (b). The edge between two instance nodes in one frame is called spatial edge~\cite{jain2016structural,vemula2017social}, which can transfer the interaction information between two traffic-agents in spatial space. The edge between the same instance in adjacent frames is the temporal edge, which is able to pass the historic information frame by frame in temporal space. The feature of the spatial edge $(A_i^t, A_j^t)$ for $A_i^t$ can be computed as $f_{ij}^t =(x_{ij}^t, y_{ij}^t, c_{ij}^t)$, where $x_{ij}^t= x_j^t - x_i^t$, $y_{ij}^t= y_j^t - y_i^t$ stands for the relative position from $A_j^t$ to $A_i^t$, $c_{ij}^t$ is an unique encoding for $(A_i^t, A_j^t)$. When traffic-agent $A_j$ considers the spatial edge, the spatial edge is represented as $(A_j^t, A_i^t)$. The feature of the temporal edge $(A_i^t, A_i^{t+1})$ is computed in the same way. 

It is normally observed that the same kind of traffic-agents have similar behavior characteristics. For example, the pedestrians have not only similar velocities but also similar reactions to other nearby traffic-agents. These similarities will be directly reflected in their trajectories. We construct a super node $C_u^t, u\in \{1,2,3\}$ for each kind of traffic-agent to learn the similarities of their trajectories and then utilize that super node to refine the prediction for instances. Fig.\ref{fig:network} (c) shows the graph in the category layer. All instances of the same type are integrated into one group and each group has an edge oriented toward the corresponding super node. After summarizing the motion similarities, the super node passes the guidance through an oriented edge to the group of instances. There are also temporal edges between the same super node in sequential frames. This category layer is specially designed for heterogeneous traffic and can make full use of the data to extract valuable information to improve the prediction results. This layer is very flexible and can be easily degenerated to situations when several categories disappear in some frames.

Finally, we get the \textit{4D Graph} for a traffic sequence with two dimensions for traffic-agents and their interactions, one dimension for time series, and one dimension for high-level categories. By this \textit{4D Graph}, we construct an information network for the entire traffic. All the information can be delivered and utilized through the nodes and edges of the graph.

\subsection{Model Architecture}
Our TrafficPredict algorithm is based on the \textit{4D Graph}, which consists of two main layers: the instance layer and the category layer. Details are given below.

\subsubsection{Instance Layer}
The instance layer aims to capture the movement pattern of instances in traffic. For each instance node $A_i$, we have an LSTM, represented as $L_i$. Because different kinds of traffic-agents have different dynamic properties and motion rules, only instances of the same type share the same parameters. There are three types of traffic-agents in our dataset: vehicles, bicycles, and pedestrians. Therefore, we have three different LSTMs for instance nodes. We also distribute LSTM $L_{ij}$ for each edge $(A_i, A_j)$ of the graph. All the spatial edges share the same parameters and all the temporal edges are classified into three types according to corresponding node type. 

For edge LSTM $L_{ij}$ at any time $t$, we embed the feature $f_{ij}^t$ into a fixed vector $e_{ij}^t$, which is used as the input to LSTM:
 \begin{align}
   & e_{ij}^t = \phi(f_{ij}^t;W_{spa}^e) ,\\
   & h_{ij}^t = LSTM(h_{ij}^{t-1};e_{ij}^t;W_{spa}^r) ,
   \end{align}
where $\phi(\cdot)$ is an embedding function, $h_{ij}^t$ is the hidden state also the output of LSTM $L_{ij}$, and $W_{spa}^e$ are the embedding weights, and $W_{spa}^r$ are LSTM cell weights, which contains the movement pattern of the instance itself. LSTMs for temporal edges $L_{ii}$ are defined in a similar way with parameters $W_{tem}^e$ and $W_{tem}^r$.

Each instance node may connect with several other instance nodes via spatial edges. However, each of the other instances has different impacts on the node's behavior. We use a soft attention mechanism~\cite{vemula2017social} to distribute various weights for all the spatial edges of one instance node： 
 \begin{equation}
    w(h_{ij}^t) = softmax(\frac{m}{\sqrt{d_e}}Dot(W_i h_{ii}^t, W_{ij} h_{ij}^t)) ,
  \end{equation}
  where $W_i$ and $W_{ij}$ are embedding weights, $Dot(\cdot)$ is the dot product, and $\frac{m}{\sqrt{d_e}}$ is a scaling factor~\cite{vaswani2017attention}. The final weights are ratios of $w(h_{ij}^t)$ to the sum. The output vector $H_i^t$ is computed as a weighted sum of $h_{ij}^t$. $H_i^t$ stands for the influence exhibited on an instance's trajectory by surrounding traffic-agents and $h_{ii}^t$ denotes the information passing by temporal edges. We concatenate them and embed the result into a fixed vector $a_{i}^t$. The node features $f_i^t$ and $a_{i}^t$ can finally concatenate with each other to feed the instance LSTM $L_i$. 
  \begin{align}
    &e_{i}^t = \phi(f_{i}^t;W_{ins}^e) ,\\
   & a_{i}^t = \phi(concat(h_{ii}^t, H_i^t);W_{ins}^a) ,\\
   & h1_{i}^t = LSTM(h2_{i}^{t-1};concat(e_{i}^t,a_{i}^t);W_{ins}^r) ,
\end{align}
where $W_{ins}^e$ and $W_{ins}^a$ are the embedding weights, $W_{ins}^r$ is the LSTM cell weight for the instance node, $h1_{i}^{t}$ is the first hidden state of the instance LSTM. $h2_{i}^{t-1}$ is the final hidden state of the instance LSTM in the last frame, which will be described in next section.

\begin{figure}
\includegraphics[width=1\columnwidth]{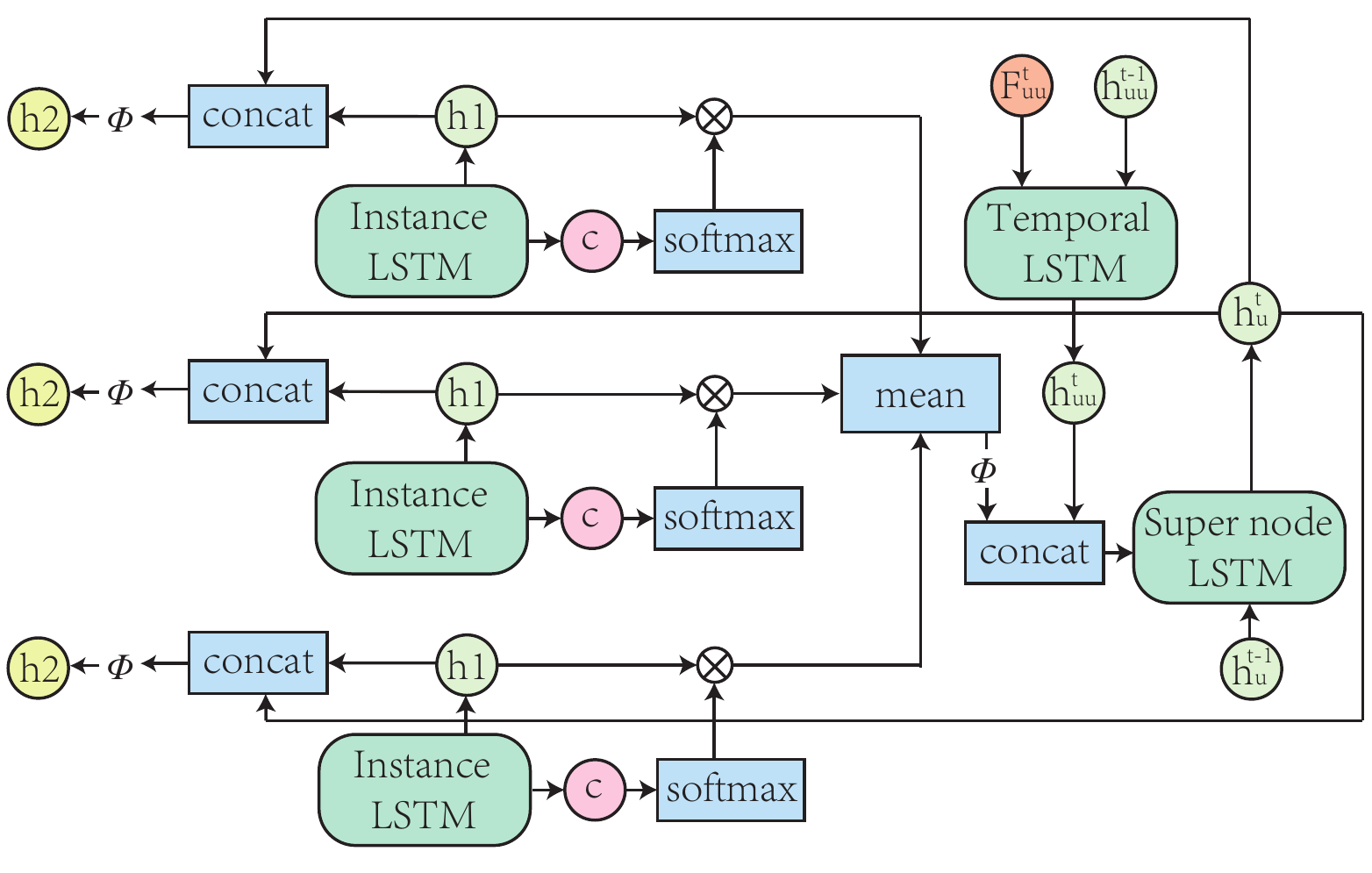}
\caption{Architecture of the network for one super node in the category layer.}
\label{fig:supernode}
\vspace{-2ex}
\end{figure}

\subsubsection{Category Layer}

Usually traffic-agents of the same category have similar dynamic properties, including the speed, acceleration, steering, etc., and similar reactions to other kinds of traffic-agents or the whole environment. If we can learn the movement patterns from the same category of instances, we can better predict trajectories for the entire instances. The category layer is based on the graph in Fig.~\ref{fig:network}(c). There are four important components: the super node for a specified category, the directed edge from a group of instances to the super node, the directed edge from the super node to instances, and the temporal edges for super nodes. 

Taking one super node with three instances as the example, the architecture in the category layer is shown in Fig.~\ref{fig:supernode}. Assume there are $n$ instances belonging to the same category in the current frame. We have already gotten the hidden state $h1$ and the cell state $c$ from the instance LSTM, which are the input for the category layer. Because the cell state $c$ contains the historical trajectory information of the instance, self-attention mechanism~\cite{vaswani2017attention} is used on $c$ by softmax operation to explore the pattern of the internal sequence. At time $t$, the movement feature $d$ for the $m$th instance in the category is captured as follows.

 \begin{equation}
    d_m^t = h1_m^t\bigotimes softmax(c_m^t) ,
  \end{equation}
 
 Then, we obtain the feature $F_u^t$ for the corresponding super node $C_u^t$ by computing the average of all the instances' movement feature of the category. 
 
  \begin{equation}
    F_u^t = \frac{1}{n} \sum_{m=1}^{n} d_m^t ,
  \end{equation}
  
  $F_u^t$ captures valid trajectory information from instances and learn the internal movement law of the category. Equation (7)-(8) show the process of transferring information on the directed edge from a group of instances to the super node.
  
The feature $ F_{uu}^t$ of the temporal edge of super node is computed by $F_{u}^t - F_{u}^{t-1}$. Take $W_{st}^e$ as embedding weights and $W_{st}^r$ as the LSTM cell weights. The LSTM of the temporal edge between the same super node in adjacent frames can be computed as follows.
  
   \begin{align}
   & e_{uu}^t = \phi(F_{uu}^t;W_{st}^e) ,\\
   & h_{uu}^t = LSTM(h_{uu}^{t-1};e_{uu}^t;W_{st}^r) ,
   \end{align}
 
 \noindent Next, we integrate the information from the group of instances and the temporal edge as the input to the super node. We embed the feature $F_u^t$ into fixed-length vectors and then concatenate with $h_{uu}^{t}$ together. The hidden state $h_{u}^t$ of super node can be gotten by follows. 
 
    \begin{align}
   & e_{u}^t = \phi(F_{u}^t;W_{sup}^e) ,\\
   & h_{u}^t = LSTM(h_{u}^{t-1};concat(e_{u}^t;h_{uu}^{t});W_{sup}^k) ,
   \end{align}

 \begin{figure}
\includegraphics[width=0.99\columnwidth]{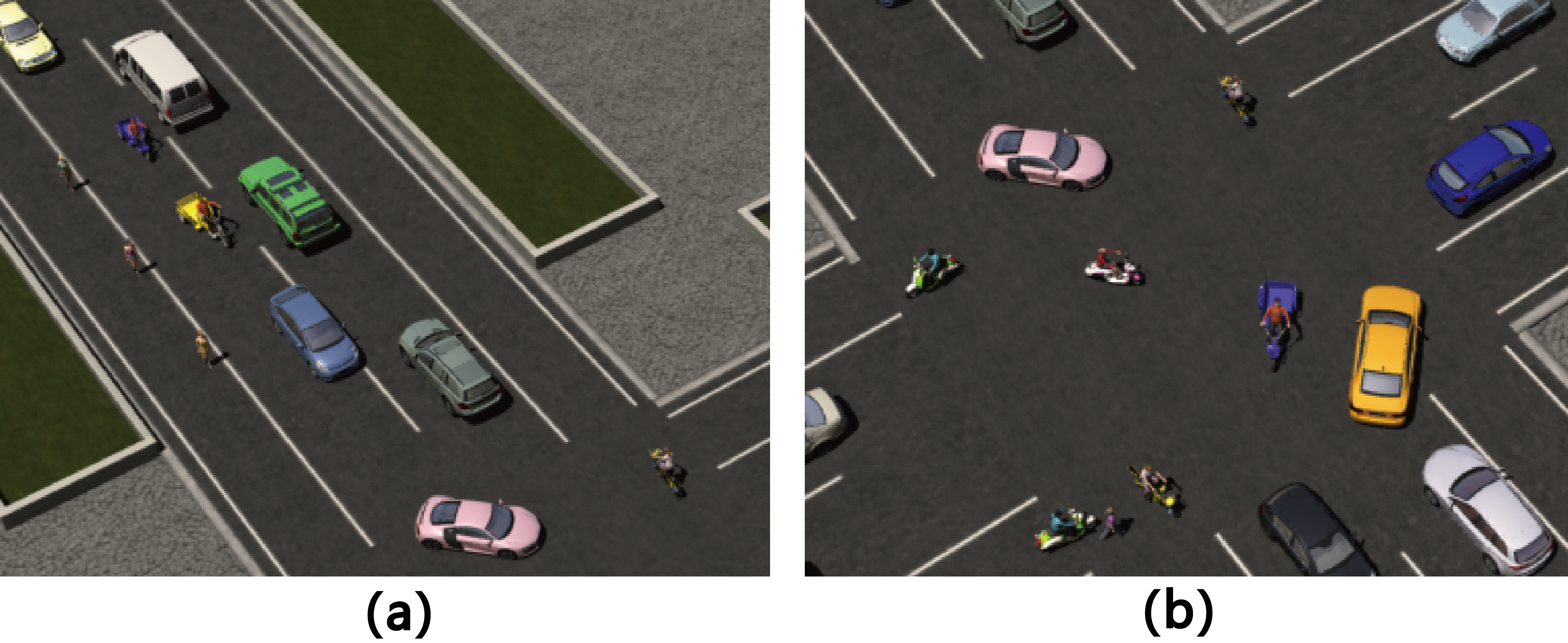}
\caption{Scenarios used for data collection: (a) Normal lanes with various traffic-agents. (b) Crossroads with different traffic-agents.}
\label{fig:data_scene}
\vspace{-3ex}
\end{figure}
   
\noindent Finally, we describe the process of transferring guidance on the directed edge from the super node to instances. For the $m$th instance in the group, the hidden state of the super node is concatenated with the first hidden state $h1_m^t$ and then embedded into a vector with the same length of $h1_m^t$. The second hidden state $h2_{m}^t$ is the final output of the instance node. 
 
 \begin{equation}
   h2_{m}^t = \phi(concat((h1_m^t;h_{u}^t);W_{s}^r)) ,
 \end{equation}
   
 \noindent where $W_{s}^r$ is the embedding weights. By the network of the category layer, we use the similarity inside the same type of instances to refine the prediction of trajectories for instances.

\subsubsection{Position estimation}
We assume the position of the traffic-agent in next frame meets a bivariate Gaussian distribution as~\cite{alahi2016social} with parameters including the mean $\mu_i^t = (\mu_x,\mu_y)_i^t$, standard deviation $\sigma_i^t = (\sigma_x,\sigma_y)_i^t$ and correlation coefficient $\rho_i^t$. The corresponding position can be represented as follows.

 \begin{equation}
  (x_i^t, y_i^t) \sim \mathcal N (\mu_i^t,\sigma_i^t,\rho_i^t) ,
 \end{equation}

The second hidden state of traffic-agents at any time is used to to predict these parameters by linear projection.
 \begin{equation}
   \lbrack \mu_i^t,\sigma_i^t,\rho_i^t\rbrack =  \phi(h2_{i}^{t-1};W_{f}) ,
 \end{equation}
The loss function is defined by the negative log-Likelihood $L_i$.
\begin{equation}
\begin{aligned}
   &L_i(W_{spa},W_{tem},W_{ins},W_{st},W_{sup},W_{s}, W_{f}) \\
   &= -\sum_{t=T_{obs}+1}^{T_{pred}} log(P(x_i^t, y_i^t | \mu_i^t,\sigma_i^t,\rho_i^t)) ,
\end{aligned}
\end{equation}

We train the model by minimizing the loss for all the trajectories in the training dataset. We jointly back-propagated through instance nodes, super nodes and spatial and temporal edges to update all the parameters to minimize the loss at each time-step.

\begin{table}
\centering
\caption{The acquisition time, total frames, total instances (count ID), average instances per frame, acquisition devices of NGSIM, KITTI (with tracklets) and our dataset.}
\label{tab:dataset_com}

\begin{tabular}{ccp{0.11\columnwidth}p{0.11\columnwidth}p{0.11\columnwidth}}
\toprule
Count& \quad &NGSIM & KITTI & Our Dataset\\
\hline
duration (min)& \quad& 45 & 22 & \textbf{155}\\
\hline
frames ($\times 10^3$)& \quad& 11.2& 13.1 &\textbf{ 93.0}\\
\hline
\multirow{3}*{\shortstack{total ($\times 10^3$)}} & pedestrian & 0 &0.09 & \textbf{16.2} \\
		~ & bicycle &  0  &  0.04 & \textbf{5.5}\\
		~ & vehicle & 2.91 & 0.93&\textbf{60.1}\\
		\hline
\multirow{3}*{average (1/f)} & pedestrian & 0 & 1.3 &\textbf{1.6} \\
		~ & bicycle &  0  &  0.24 & \textbf{1.9} \\
		~ & vehicle & \textbf{845} & 3.4 & 12.9\\
		\hline
\multirow{3}*{device} & camera & yes &yes &yes \\
		~ & lidar &  no  &  yes & yes \\
		~ & GPS &no & yes & yes \\
\bottomrule
\end{tabular}
\end{table}

\begin{table*}
\centering
\caption{The average displacement error and the final displacement error of the prior  methods (ED, SL, SA) and variants of our method (TP) on our new dataset. For each evaluation metric, we show the values on pedestrians, bicycles, vehicles, and all the traffic-agents. We set the observation time as 2 seconds and the prediction time as 3 seconds for these measurements.}
\label{tab:comparison_error}
\begin{tabular}{c|p{1.5cm}<{\centering}|p{1.5cm}<{\centering}|p{1.5cm}<{\centering}|p{1.5cm}<{\centering}|c|c|c}
\toprule
Metric& Methods & ED & SL & SA & TP-NoCL & TP-NoSA & TrafficPredict\\
\hline
\multirow{4}*{Avg. disp. error} & pedestrian & 0.121 & 0.135 & 0.112 & 0.125 & 0.118 & \textbf{0.091} \\
		~ & bicycle &  0.112  &  0.142 & 0.111 & 0.115 & 0.110 & \textbf{0.083}\\
		~ & vehicle & 0.122 & 0.147 &0.108 & 0.101 & 0.096 & \textbf{0.080}\\
		~ & total & 0.120 & 0.145&0.110 & 0.113 & 0.108 & \textbf{0.085}\\
		\hline
\multirow{4}*{Final disp. error} & pedestrian & 0.255 & 0.173 & 0.160 & 0.188 & 0.178 & \textbf{0.150}\\
		~ & bicycle &  0.190  &  0.184 & 0.170 & 0.193 & 0.169 & \textbf{0.139}\\
		~ & vehicle & 0.195 & 0.202 &0.189 & 0.172 & 0.150 & \textbf{0.131}\\
		~ & total & 0.214 & 0.198&0.178 & 0.187 & 0.165 & \textbf{0.141}\\

\bottomrule

\end{tabular}
\end{table*}

\section{Experiments}

\subsection{Dataset}
We use Apollo acquisition car~\cite{baiduapollo} to collect traffic data, including camera-based images and LiDAR-based point clouds, and generate trajectories by detection and tracking. 

Our new dataset is a large-scale dataset for urban streets, which focuses on trajectories of heterogeneous traffic-agents for planning, prediction and simulation tasks. Our acquisition car runs in urban areas in rush hours in those scenarios shown in Fig.~\ref{fig:data_scene}. The data is generated from a variety of sensors, including LiDAR (Velodyne HDL-64E S3), radar (Continental ARS408-21), camera, high definition maps and a localization system at 10HZ. We provide camera images and trajectory files in the dataset. The perception output information includes the timestamp, and the traffic-agent's ID, category, position, velocity, heading angle, and bounding polygon. The  dataset  includes RGB videos with 100K $1920\times1080$ images and around 1000km trajectories for all kinds of moving traffic agents. A comparison of NGSIM, KITTI (with tracklets), and our dataset is shown in Table.~\ref{tab:dataset_com}. Because NGSIM has a very large, top-down view, it has a large number of vehicles per frame. In this paper, each period of sequential sequences of the dataset was isometrically normalized for experiments. 
Our new dataset has been released over the WWW~\cite{apolloscapeTrajectory}.

\subsection{Evaluation Metrics and Baselines}

We use the following metrics~\cite{pellegrini2009you,vemula2017social} to measure the performance of algorithms used for predicting the trajectories of traffic-agents. 

\begin{enumerate}[1.]
\item \textit{Average displacement error}: The mean Euclidean distance over all the predicted positions and real positions during the prediction time.
\item \textit{Final displacement error}: The mean Euclidean distance between the final predicted positions and the corresponding true locations. 
\end{enumerate}

We compare our approach with these methods below:
\begin{itemize}
\item {\textit{RNN ED (ED)}: An RNN encoder-decoder model, which is widely used in motion and trajectory prediction for vehicles.}
\item {\textit{Social LSTM (SL)}: An LSTM-based network with social pooling of hidden states~\cite{alahi2016social}. The model performs better than traditional methods, including the linear model, the Social force model, and Interacting Gaussian Processes.}
\item {\textit{Social Attention (SA)}: An attention-based S-RNN architecture~\cite{vemula2017social}, which learn the relative influence in the crowd and predict pedestrian trajectories. }
\item {\textit{TrafficPredict-NoCL (TP-NoCL)}: The proposed method without the category layer.}
\item {\textit{TrafficPredict-NoSA (TP-NoSA)}: The proposed method without the self-attention mechanism of the category layer.}
\end{itemize}

\subsection{Implementation Details}
In our evaluation benchmarks, the dimension of hidden state of spatial and temporal edge cell is set to 128 and that of node cell is 64 (for both instance layer and category layer). We also apply the fixed input dimension of 64 and attention layer of 64. During training, Adam~\cite{adam} optimization is applied with $\beta_1$=0.9 and $\beta_2$=0.999. Learning rate is scheduled as 0.001 and a staircase weight decay is applied. The model is trained on a single Tesla K40 GPU with a batch size of 8. For the training stability, we clip the gradients with the range -10 to 10. During  the computation of predicted trajectories, we observe trajectories of 2 seconds and predict the future trajectories in next 3 seconds.

\subsection{Analysis}

\begin{figure*}
\includegraphics[width=2.1\columnwidth]{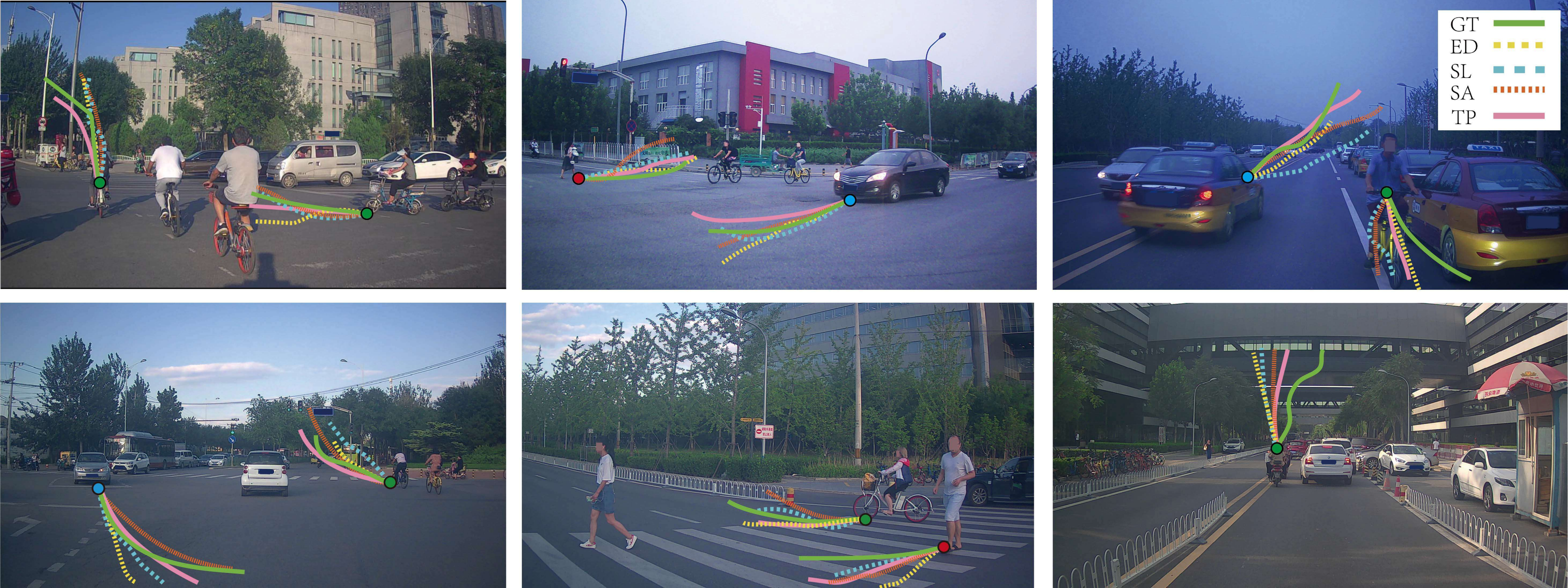}
\caption{Illustration of our TrafficPredict (TP) method on camera-based images. There are six scenarios with different road conditions and traffic situations. We only show the trajectories of several instances in each scenario. The ground truth (GT) is drawn in green and the prediction results of other  methods (ED,SL,SA) are shown with different dashed lines. The prediction trajectories of our TP algorithm (pink lines) are the closest to ground truth in most of the cases.}
\label{fig:result1}
\vspace{-3ex}
\end{figure*}

\begin{figure}
\includegraphics[width=1\columnwidth]{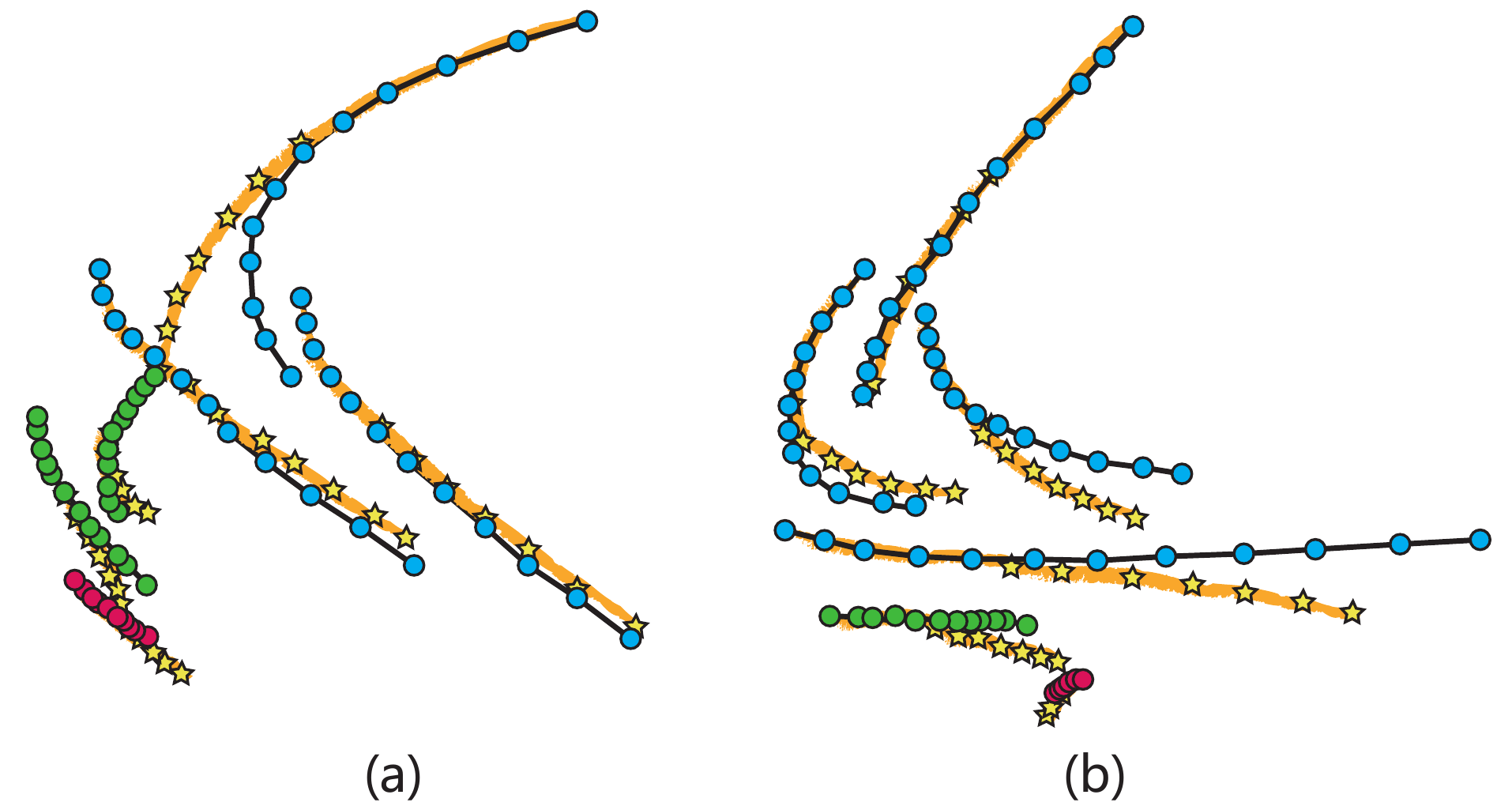}
\caption{Illustration of some prediction results by our method. The ground truth of trajectories of vehicles, bicycles and pedestrians are drawn by blue, green and red respectively. Predicted locations are all represented by yellow stars. For each instance, first five discrete points are observed positions, but there are some overlaps in the illustration of pedestrian trajectories. }
\label{fig:result2}
\vspace{-4ex}
\end{figure}

The performance of all the prior methods and our algorithm on heterogeneous traffic datasets is shown in Table.~\ref{tab:comparison_error}. We compute the average displacement error and the final displacement error for all the instances and we also count the error for pedestrians, bicycles and vehicles, respectively. The social attention (SA) model considers the spatial relations of instances and has smaller error than RNN ED and Social LSTM. Our method without category layer (TP-NoCL) not only considers the interactions between instances but also distinguishes between instances by using different LSTMs. Its error is similar to SA. By adding the category layer without self attention, the prediction results of TP-NoSA are more accurate in terms of both metrics. The accuracy improvement becomes is more evident after we use the self-attention mechanism in the design of category layer. Our algorithm, TrafficPredict, performs better in terms of all the metrics with about 20\% improvement of accuracy. It means the category layer has learned the inbuilt movement patterns for traffic-agents of the same type and provides good guidance for prediction. The combination of the instance layer and the category layer makes our algorithm more applicable in heterogeneous traffic conditions.

We illustrate some prediction results on corresponding 2D images in Fig.~\ref{fig:result1}. The scenario in the image captured by the front-facing camera does not show the entire scenario. However, it is more intrinsic to project the trajectory results on the image. In most  heterogeneous traffic scenarios, our algorithm computes a reasonably accurate predicted trajectory and is close to the ground truth. If we have prior trajectories over a longer duration, the prediction accuracy increases.

When traffic-agents are moving on straight lanes, it is easy to predict their trajectories because almost all the traffic-agents are moving in straight direction. It is more challenging to provide accurate prediction in cross roads, as the agents are turning. Fig.~\ref{fig:result1} shows 2D experimental results of two sequences in cross areas. There are some overlaps on trajectories. 
In these scenarios, there are many curves with high curvature because of left turn. Our algorithm can compute accurate predicted trajectories in these cases.

\section{Conclusion}
In this paper, we have presented a novel LSTM-based algorithm, TrafficPredict, for predicting trajectories for heterogeneous traffic-agents in urban environment.  
We use a instance layer to capture the trajectories and interactions for instances and use a category layer to summarize the similarities of movement pattern of instances belong to the same type and guide the prediction algorithm. All the information in spatial and temporal space can be leveraged and transferred in our designed \textit{4D Graph}. Our method outperforms previous state-of-the-art approaches in improving the accuracy of trajectory prediction on our new collected dataset for heterogeneous traffic. We have evaluated our algorithm in traffic datasets corresponding to urban dense scenarios and observe good accuracy. Our algorithm is realtime and makes no assumption about the traffic conditions or the number of agents.

Our approach has some limitations. Its accuracy varies based on traffic conditions and the duration of past trajectories. In the future, we will consider more constraints, like the lane direction, the traffic signals and traffic rules, etc. to further improve the accuracy of trajectory prediction. Furthermore, we would like to evaluate the performance in more dense scenarios.

\section{Acknowledgement}
Dinesh Manocha is supported in part  by ARO Contract W911NF16-1-0085, and Intel. We appreciate all the people who offered help for collecting the dataset. 

\appendix

\small
\bibliographystyle{named}
\bibliography{aaai}

\end{document}